\documentclass{article}

\usepackage[final]{corl_2020} % Uncomment for the camera-ready ``final'' version.
\usepackage{graphicx}
\usepackage{amsmath}
\usepackage{amssymb}
\usepackage{subcaption}
\usepackage{babel,blindtext}
\usepackage{booktabs}
\usepackage{multirow}
\usepackage{bm}
\usepackage{xspace}

\def\eg{\emph{e.g.}}

\def\etal{\emph{et al.}}
\makeatother

\title{CLOUD: Contrastive Learning of Unsupervised Dynamics}

\author{
  Jianren Wang\thanks{Contribute equally.}\\
  Carnegie Mellon University \\ 
  \texttt{jianrenw@andrew.cmu.edu} \\
   \And
   Yujie Lu\textsuperscript{\textasteriskcentered}\\
   Tencent \\
   \texttt{yujielu10@gmail.com} \\
   \AND
   Hang Zhao \\
   Tsinghua University\\ Shanghai Qizhi Institute \\
   \texttt{zhaohang0124@gmail.com} \\
}

\begin{document}
\maketitle

%===============================================================================

\begin{abstract}

Developing agents that can perform complex control tasks from high dimensional observations such as pixels is challenging due to difficulties in learning dynamics efficiently. In this work, we propose to learn forward and inverse dynamics in a fully unsupervised manner via contrastive estimation. Specifically, we train a forward dynamics model and an inverse dynamics model in the feature space of states and actions with data collected from random exploration. Unlike most existing deterministic models, our energy-based model takes into account the stochastic nature of agent-environment interactions. We demonstrate the efficacy of our approach across a variety of tasks including goal-directed planning and imitation from observations. Project videos and code are at \url{https://jianrenw.github.io/cloud/}. 

\end{abstract}


\section{Introduction}

Modeling dynamics in the physical world is an essential and fundamental problem in the robotics applications, such as manipulation~\cite{de1988compliant, karayiannidis2016adaptive}, planning~\cite{kurutach2018learning, hafner2019learning,pathakICLR18zeroshot}, and model-based reinforcement learning~\cite{8463189, kaiser2019model}. To perform complex control tasks in an unknown environment, an agent needs to learn the dynamics from experience~\cite{nair2017combining,agrawal2016learning, kumar2016optimal}. However, developing agents that can learn dynamics directly from high dimensional observations such as pixels is known to be challenging~\cite{lake2017building, kaiser2019model, laskin_srinivas2020curl}.  

There are three key challenges. First, learning informative representations of the states from raw images is difficult. Second, the dynamics of real-world objects are always complex and nonlinear, especially for deformable objects. Third, modeling the stochasticity of the agent-environment system is extremely challenging. The stochasticity mainly comes from two aspects: the noise in the agent's actuation and the inherent uncertainty in the environment, both of which cannot be ignored. 
Many recent works have been proposed to tackle these challenges. A direct approach is to learn complex dynamics models from the pixel space~\cite{chen2016infogan, ebert2018visual}. Kurutach~\etal~\cite{kurutach2018learning} suggested learning a generative model of sequential observations, where the generative process is induced by a transition in a low-dimensional planning model, and additional noise. Their work can model the stochastic transition function. However, making predictions in raw sensory space is not only hard but is also irrelevant to the agent's goals, \textit{e.g.} a model needs to capture appearance changes in order to make photo-realistic predictions, which makes it hard to generalize across physically similar but visually distinct environments. One possible solution is to predict those changes in the space that affect or be affected by the agent, and ignore rest of the unrelated dynamics. For example, instead of making predictions in the pixel space, we could transform the sensory input into a feature space where only the information relevant to the agent actions are represented. Agrawal~\etal~\cite{agrawal2016learning} proposed to jointly train forward and inverse dynamics models, where a forward model predicts the next state from the current state and action, and an inverse model predicts the action given the initial and target states. In the joint training, the inverse model objective provides supervision for transforming image pixels into an abstract feature space, while the forward model can predict in it. The inverse model alleviates the need for the forward model to make predictions in the raw pixel space, and the forward model in turn regularizes the feature space for the inverse model. This simple strategy has been adopted by many works~\cite{nair2017combining,pathak2017curiosity}. However, these methods can hardly learn predictive models beyond noise-free environments. Despite several methods to build stochastic models in low-dimensional state space~\cite{chua2018deep,houthooft2016vime}, scaling it to high dimensional inputs (\eg, images) still remains challenging.

In this paper, we introduce a novel method, named CLOUD, that uses contrastive learning~\cite{chen2020simple} to handle all the above challenges. Inspired by~\cite{agrawal2016learning}, we jointly train a forward dynamics model, an inverse dynamics model, a state representation model, and an action representation model with contrastive objectives. The forward dynamics model is trained to maximize the agreement between the predicted and the observed next state representations; the inverse dynamics model is trained to maximize the agreement between the predicted and the ground truth action representations. Our proposed approach offers three unique advantages. First, contrastive learning only measures the compatibility between prediction and observation, which can handle the stochasticity by nature~\cite{lecun2006tutorial}. Second, by introducing an action representation, our method can generalize over large, finite action sets by allowing the agent to infer the outcomes of actions
similar to actions previously taken~\cite{chandak2019learning}. Third, building upon SimCLR~\cite{chen2020simple}, our proposed method is data-efficient and can be trained in a fully unsupervised manner. 

To summarize, 1) we propose a general framework to learn a forward dynamics model, an inverse dynamics model, a state representation model, and an action representation model jointly; 2) using contrastive estimation, our proposed method can handle the stochasticity of the agent-environment system naturally; 3) we demonstrate the efficacy of our approach across a variety of tasks including goal-directed planning and imitation from observations.

\section{Related Works}

\paragraph{Contrastive Learning}

Contrastive Learning is a framework to learn representations that obey similarity constraints
in a dataset typically organized by similar and dissimilar pairs. There has been a large number of prior works. Hadsell et al.~\cite{hadsell2006dimensionality} first propose to learn representations by contrasting positive pairs against negative pairs. Wu et al.~\cite{wu2018unsupervised} propose to use a memory bank to store the instance class representation vector, which is adopted and extended by several recent papers~\cite{ye2019unsupervised, tian2019contrastive}. Other work explores the use of in-batch samples for negative sampling instead of a memory bank~\cite{doersch2017multi,ye2019unsupervised, ji2019invariant}. Recently, SimCLR~\cite{chen2020simple} and MoCo~\cite{he2020momentum, chen2020mocov2} achieved state-of-the-art results in self-supervised visual representations learning, bridging the gap with supervised learning.

\paragraph{Learning Dynamics}

Modelling dynamics is a long-standing problem in both robotics and artificial intelligence. One line of work is to estimate physical properties directly from their appearance and motion, which is known as intuitive physics~\cite{wu2015galileo, wu2017learning, watters2017visual, ye2018interpretable, ehrhardt2019taking, fragkiadaki2015learning}. These models build upon an explicit physical model, i.e., a model parameterized by physical properties such as mass and force. This enables generalization to new scenarios, but also limits their practical usage: annotations on physical parameters in real-world applications are expensive and challenging to obtain. An alternative line of work is to learn object representations without explicit modeling of physical properties, but in a self-supervised way through robot interactions. Byravan et al.~\cite{byravan2017se3} propose to use deep networks to approximate rigid object motion. Agrawal et al.~\cite{agrawal2016learning} suggest encoding physical properties in latent representations that can be decoded through forward and inverse dynamics models. A few follow-ups have extended these models for rope manipulation~\cite{nair2017combining}, pushing via transfer learning~\cite{7989249}, 
and planning~\cite{kurutach2018learning, hafner2019learning}. A concurrent work from Yan etal.~\cite{yan2020learning} has also demonstrated the effectiveness of using contrastive estimation to learn predictive representations. Different from them, we focus on forward and backward dynamics reasoning under stochastic environment.

\paragraph{Imitation from Observations}

Increasingly, works have aspired to learn from observation alone without utilizing expert actions~\cite{edwards2019imitating}. \textit{e.g.}, Liu et al.~\cite{liu2018imitation} propose to learn to imitate from videos without actions and translates from one context to another. Ho et al.~\cite{ho2016generative} propose to learn features for a reward signal that is later used for reinforcement learning. Few recent works aim to learn inverse dynamics in a self-supervised manner, then given a task, attempt it zero-shot~\cite{pathakICLR18zeroshot, torabi2018behavioral, edwards2019imitating}.

\section{Method}

\begin{figure}
  \includegraphics[width=\textwidth]{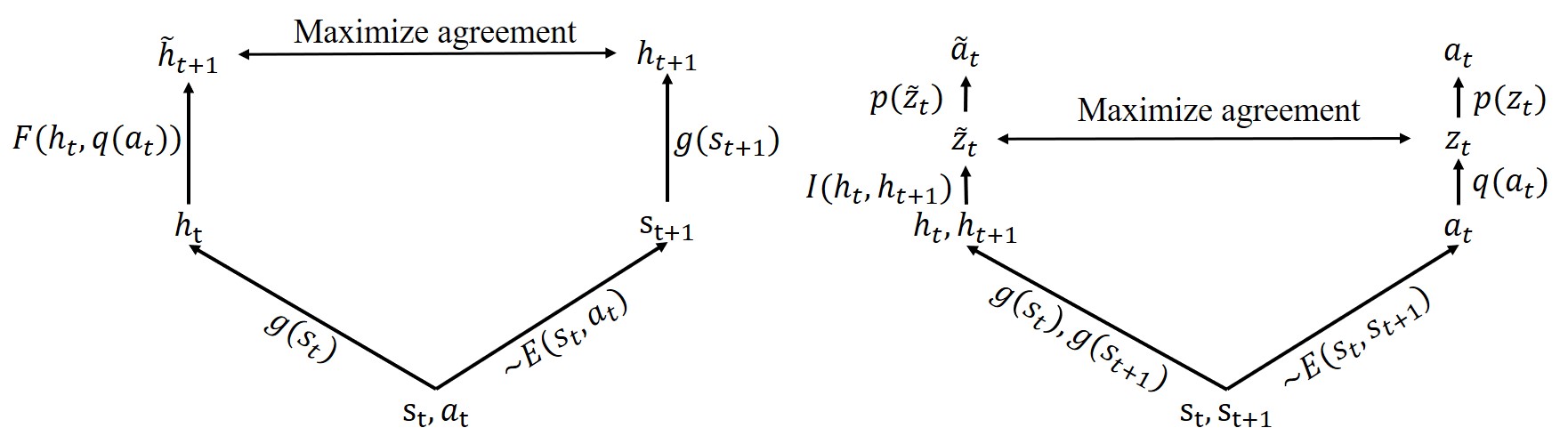}
  \caption{CLOUD Architecture: Our framework consists of four learnable functions, including a forward dynamics model $F(\cdot)$, an inverse dynamics model $I(\cdot)$, a state representation model $g(\cdot)$ and an action representation model $q(\cdot)$. We propose to learn these four models jointly via contrastive estimation.}
  \label{fig:pipeline}
\end{figure}

Our framework consists of four learnable functions, including a forward dynamics model $F(\cdot)$, an inverse dynamics model $I(\cdot)$, a state representation model $g(\cdot)$ and an action representation model $q(\cdot)$. We propose to learn these four models jointly via contrastive estimation. We begin by discussing the state representation model and action representation model. Following that, we discuss the forward dynamics model and inverse dynamic model. Finally, we discuss how to use contrastive estimation to jointly optimize these four models in an unsupervised manner. The notation is as following: $s_t, a_t$ are the world state and action applied time step $t$, $s_{t+1}$ is the world state at time step $t+1$. $E$ represents the stochastic environment, where $s_{t+1}$ can be sampled from $s_{t+1} \sim E(s_t,a_t)$. See Figure~\ref{fig:pipeline} for an illustration of the formulation. 

\subsection{State and Action Representation Models}

The benefits of capturing the structure in the underlying state space is a well understood and widely used concept in robotics. Given a set of states $\{s\} \subseteq \mathcal{S}$ (where the states are raw sensory inputs), the goal of state representation model $g(\cdot)$ is to encode high-dimensional images $s$ into informative representations $h$, which is mathematically described as $h = g (s)$. State representations allow the policy to generalize across states. Ideally, the value comparisons between these features should correlate well with true distances between these states. Naturally, such features would be useful later for planning. 

Similarly, there often exists additional structure in the space of actions that can be leveraged. Given a set of actions $\{a\} \subseteq \mathcal{A}$, we introduce an action representation model $q(\cdot)$ to encode actions $a$ into embeddings $z$, which is mathematically described as $z = q (a)$. We then propose to use an action decoder $p(\cdot)$ that deterministically maps this representation to the action, which is denoted as $a = p (z)$. 

\subsection{Forward and Inverse Dynamics Models}

Forward dynamics models and inverse dynamics models have been studied for a long time. Let $\{(s_t, a_t, s_{t+1})\}$ represent a set of state-action-state tuples. A model that predicts the state of the next timestep $\tilde{s}_{t+1}$ given current state and action $s_t, a_t$ is known as a forward dynamics model $F(\cdot)$, and is mathematically described as $\tilde{s}_{t+1} = F(s_t, a_t)$. Instead of directly learning a dynamics model through pixel space, we consider to make predictions in a learned latent space. Because making predictions in raw sensory space is hard and can always be distracted by irrelevant information~\cite{kurutach2018learning}. Thus, our forward dynamics model can be mathematically described in Equation~\ref{eq:forward} below:
\begin{equation}
    \tilde{h}_{t+1} = F(h_t, z_t),
\label{eq:forward}
\end{equation}
where $h_t = g (s_t), z_t = q (a_t)$.

Oe other hand, a model that predicts the action $a_t$ that relates a pair of input states $(s_t, s_{t+1})$ is called an inverse dynamics model $I(\cdot)$. Similarly, we consider to make predictions in a learned action space, which is described as following:
\begin{equation}
    \tilde{z}_{t} = I(h_t, h_{t+1}),
\label{eq:inverse}
\end{equation}
where $h_t = g (s_t), h_{t+1} = g (s_{t+1})$.

\subsection{Contrastive Estimation}

Many works~\cite{nair2017combining, agrawal2016learning, pathakICLR18zeroshot} directly optimize the above mentioned models by forcing $h_{t+1} = \tilde{h}_{t+1}$ and $z_{t} = \tilde{z}_{t}$. However, by forcing predicted representations equal to observed future representations, these methods also assume the transition to be deterministic, which is always not true in the real world. 
% \HZ{is our method free of this assumption? If yes, more discussion is necessary}. 
The real environment is always stochastic (\textit{e.g.} coin toss), where deterministic functions can only predict the average. 

On the other hand, contrastive estimation is an energy-based model. Instead of setting the cost function to be zero only when the prediction and the observation are the same, energy-based model assigns low cost to all compatible prediction-observation pairs. Thus, constrastive estimation can handle the stochasticity by its nature~\cite{lecun2006tutorial}. Inspired by recent contrastive learning algorithms~\cite{chen2020simple}, we propose to train these models by maximizing agreement between predicted and real representations via a contrastive loss. We randomly sample a minibatch of N state-action-state tuples $\{(s^i_t, a^i_t, s^i_{t+1})\}$. For forward dynamics model, a prediction $\tilde{h}^i_{t+1}$ and real representation $h^i_{t+1}$ from the same tuple is defined as positive example. Following SimCLR~\cite{chen2020simple}, we treat the other $2(N - 1)$ real representation $(h^j_t, h^j_{t+1}) | j \neq i$ within a minibatch as negative examples. We use cosine similarity to denote the distance between two representation $(u, v)$, that is $\texttt{sim}(u,v) = \mathbf{u}^T \cdot \mathbf{v}/||\mathbf{u}|| \cdot ||\mathbf{v}||$. The loss function for a positive pair of examples $(\tilde{h}^i_{t+1}, h^i_{t+1})$ is defined as:

\begin{equation}
l_F = -\text{log} \frac{\text{exp}(\texttt{sim}(\tilde{h}^i_{t+1}, h^i_{t+1})/\tau)}{\sum_{\substack{j=1 \\ j\neq i}}^{N} \text{exp}(\texttt{sim}(\tilde{h}^i_{t+1}, h^j_t)/\tau) + \sum_{\substack{j=1 \\ j\neq i}}^{N} \text{exp}(\texttt{sim}(\tilde{h}^i_{t+1}, h^j_{t+1})/\tau)},
\label{eq:forward_contrastive}
\end{equation}
where $\tau$ denotes a temperature parameter that is empirically chosen as $0.1$.

Similarly, for inverse dynamics model, the loss function for a positive pair of examples $(\tilde{z}^i_t, z^i_t)$ is defined as:

\begin{equation}
l_I = -\text{log} \frac{\text{exp}(\texttt{sim}(\tilde{z}^i_{t}, z^i_{t})/\tau)}{\sum_{\substack{j=1 \\ j\neq i}}^{N} \text{exp}(\texttt{sim}(\tilde{z}^i_{t}, z^j_t)/\tau) + \sum_{\substack{j=1 \\ j\neq i}}^{N} \text{exp}(\texttt{sim}(\tilde{z}^i_{t}, z^j_{t+1})/\tau)}.
\label{eq:inverse_contrastive}
\end{equation} 

During joint training, the total loss is computed across all positive pairs in a mini-batch.

\section{Experiments}
We evaluate our proposed method in comparison with classical baseline methods on two tasks, goal-directed planning and imitation from observations on rope manipulation. The details of the datasets and other experimental settings are described below.

\subsection{Representation Visualization}

To understand the representation models, we first propose to visualize the state representations learned by our model using t-SNE~\cite{maaten2008visualizing}. With t-SNE visualizations there tends to be many overlapping points in the 2D space,  which increases the difficulty of viewing overlapped state examples. Therefore, we quantize t-SNE points into a 2D grid with $40\times20$ interface, using RasterFairy~\cite{mario2015RasterFairy}.

\begin{figure}
\centering\includegraphics[scale=0.15]{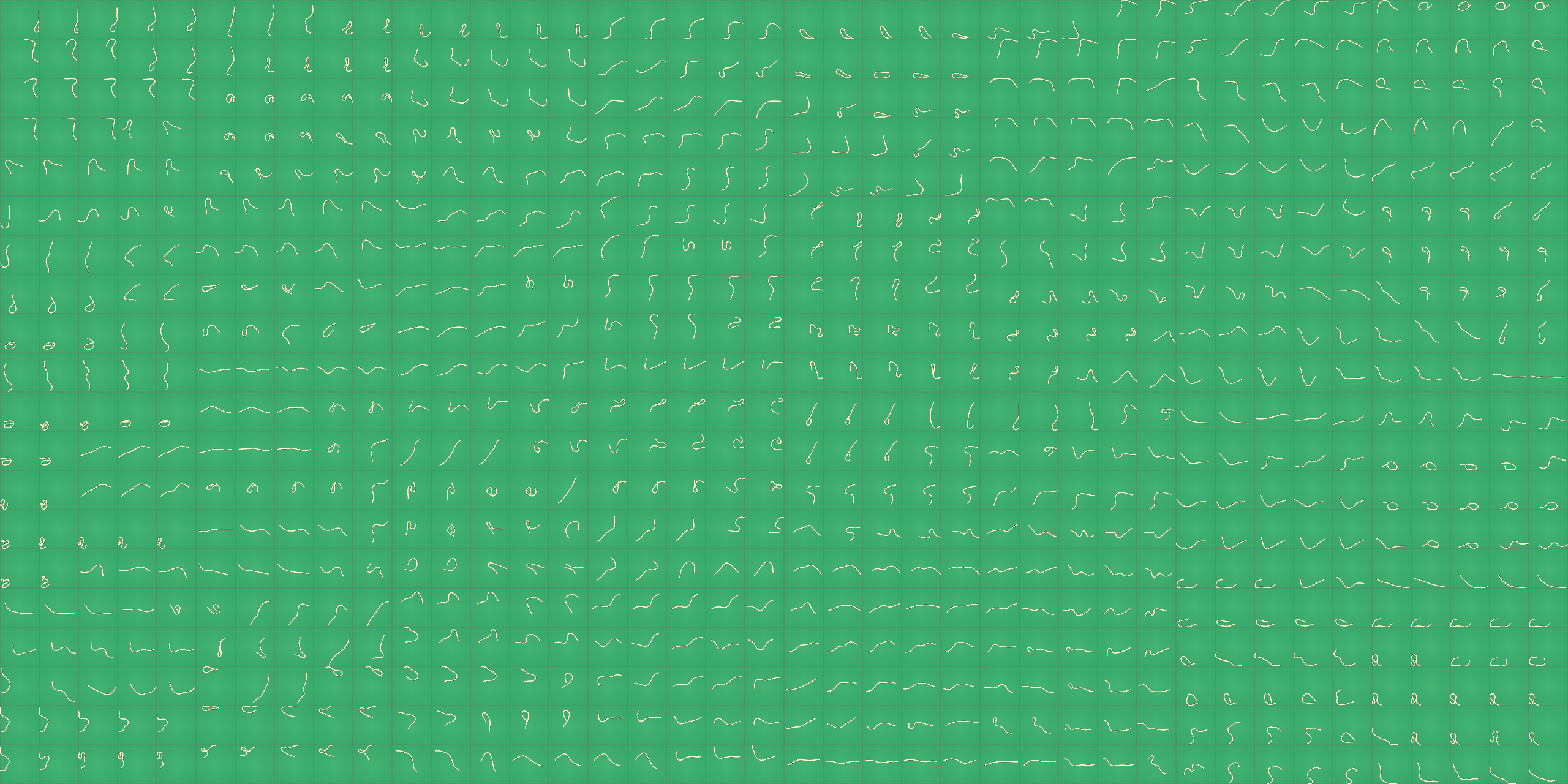}
  \caption{\textbf{t-SNE Visualization of State Representation.} The grid interface is $40\times20$, fit with image examples from validation set.
}
  \label{fig:tsne-state}
  \vspace{1em}
\end{figure}

In Figure~\ref{fig:tsne-state}, we show the state representations from validation set which are not seen during training. Notice that similar configurations of the rope appear near each other, indicating the learned feature space meaningfully organizes variation in rope shape.

\subsection{Goal-Directed Planning}

In goal-directed planning, an agent is required to generate a plausible sequence of states that transition a dynamical system from its current configuration to the desired goal state. Given the learned representations and dynamics models, we propose to plan with a simple version of goal-directed planning (MPC). We first sample several actions and then feed them to the forward dynamics model. Finally, we choose the action that leads to the largest cosine similarity between predicted state representation and goal state representation. The procedure is repeated iteratively for 20 steps.

\paragraph{Dataset and Settings.}
\label{sec:data_goal}
To simulate rope, we use the DeepMind Control Suite~\cite{tassa2018deepmind} with MuJoCo physics engine~\cite{todorov2012mujoco}. The rope is represented by 25 geoms in simulation with a four-dimensional action space $(x_1,y_1,x_2,y_2)$: the first two dimensions $(x_1,y_1)$ denote the pick point on the rope, and the last two dimensions $(x_2,y_2)$ are the drop location, both of which are in the pixel coordinates of the input RGB image. In order to evaluate the performance of CLOUD under a stochastic environment, we add Gaussian noise to each geom after actions are executed. We use an overhead camera that renders RGB images as input observations for training our model.

% To evaluate the ability of our approach to develop an agent that could perform complex control tasks by receiving high dimensional observations such as pixels effectively, we use deformable objects for manipulation in our experiments. Both of the tasks are performed in our custom rope environment extended from DeepMind Control Suite with MuJoCo as physics engine. , which is a challenging problem in reinforcement learning. It is known that many optimal goal-directed planning still struggle with stochastic environment. In this work we present a model jointly learning four learnable components via contrastive estimation to tackle this problem naturally. As the basic rope simulation lacks stochasticity, we construct another custom environment to evaluate the performance on this particular task. % In order to construct such a stochastic environment, we integrate the Gaussian noise spatially over the original states which represent the locations of all the composite objects in simulation. 

By randomly perturbing the rope, our data is collected in a completely unsupervised manner. The interaction of the agent with rope is uniformly sampled but constrained to the observed field to avoid redundant data. We collect 10k trajectories of length 20 (200k samples), which are further split into 150k training samples and 50k testing samples.

% As for the data collection process, we utilize the reinforcement learning platform illustrated above to collect a large amount of data in a self-supervised manner by executing random available actions on the observed rope at each step. The interaction of the agent with rope is uniformly sampled but constrained to the observed field to avoid redundant data. Both the actions and states saved for each run are then split into training and testing dataset.

% \JW{I am wrong here. } 

We evaluate the performance of goal-directed planning on two types of goal states: shaped and straight. Following~\cite{agrawal2016learning}, we manually pick a set of complex rope shapes for an agent to reach, including "C", "L", "S" and "knot", which are denoted as "shaped goal state". For straight goal state, the agent is supposed to straighten the rope from a given initial configuration.  

The performance is quantitatively measured by calculating Euclidean distance between two sets of geoms from achieved and goal states correspondingly, which attempt to capture the deviation between desired goals and agent achieved final states. A successful manipulation is judged by whether the mean distance error is below the threshold (set according to human observation, we use 4 of each geom in pixel space) at each run to calculate the success rate.

% To quantitatively evaluate the performance of the agents on goal-directed planning in rope manipulation environment, we use objective metrics including success rate and distance error, which attempt to capture the deviation between desired goals and agent achieved states. A successful manipulation is judged by whether the average distance error is below the threshold (set according to human investigation) at each run, which is then summed up to calculate the average success rate of each agent based on different policy for all the tasks. 

% \JW{what does this mean?}
% During inference, we compare our proposed model to the classical baselines according to the objective metrics. We re-implement the baselines and train them in the same experimental settings for a fair comparison. In order to compare the performance of these models on predicting the next state from the current state and action, we computed overall distance error of the two set of matching components of the deformable rope object between demonstrations and state sequences produced by the agents.

\paragraph{Training Details.}
\label{network}
% During training of our proposed method, the states and actions are embedded through our learnable state and action representation component respectively, \JW{Maybe we don't want to mention this?} and then concatenated with the Gaussian noise input before fed into the 2-layer MLP(multilayer perception).  The incorporation with the Gaussian noise over our input is meant to help the model learn how to deal with the stochasticity in the environment. 

For state representation model $g(\cdot)$, we use a series of 2D convolutions to extract useful features from $64\times64$ raw RGB images. The output is then flattened and fed into a linear layer to produce low-dimensional embeddings of the state representations $h$ in $\mathbf{R^{16}}$. For action representation model $g(\cdot)$, we use a 4-layer MLP to extract useful features from actions and output 16-dimension action embeddings. The forward model is a 4-layer MLP which takes a state representation, and an action representation as input and then outputs the representation of the next state. The inverse model is a 4-layer MLP which takes two state representations as input and then outputs the corresponding action representation.

% The experiment results show a little boost to the performance, when added the noise to our model during training period. 
We use the Adam optimizer~\cite{kingma2014adam} for training the network with a batch size of 128 and a learning rate of 1e-3 with a weight decay of 1e-6. We train the network for 30 epochs and report the average success rate for evaluation.

% \JW{Explain what is the action embedding in CLOUD(F).}
% To understand the effects of our learning technique, we ablate several variants of our proposed model, including a single forward dynamics model, named as CLOUD(F), and a single inverse dynamics model, named as CLOUD(I). We perform comparisons to review the importance of each component for our complete model quantitatively and qualitatively during evaluation.
 
% old \subsection{Models for Comparison}

\paragraph{Results.} % introduction of all baselines
Table~\ref{tab:mpc} shows the success rate for various approaches for goal-directed planning. We evaluate two different variants of our method:

% Our goal-directed planning experiments aim to have the agent predict the next state given the current state and action. We compare our method with several classical baselines on the MPC task, including Auto-encoder, PlaNet, Predictive Model and a variant of our complete model. The details of all the models in comparison are described below. 

\begin{itemize}
    \item \textit{CLOUD (F)}: This method refers to a variant of CLOUD without the inverse dynamics model, where we feed raw actions instead of action embeddings to the forward dynamics model. The purpose of this variant is to particularly ablate the benefit of our inverse dynamics model.
    \item \textit{CLOUD (FI)}:  Our complete model composes of all the four components, including a forward dynamics component, an inverse dynamics component, a state representation component and an action representation component. These components are trained jointly via contrastive estimation in an unsupervised manner.
\end{itemize}

We compare our results to the following baselines:

\begin{itemize}
    \item \textit{Auto-encoder}: We train a simple autoencoder to minimize the pixel distance between reconstructed and actual images~\cite{lange2010autoencoder}. The latent embedding is then used for MPC during planning.
    \item \textit{PlaNet}: We train PlaNet~\cite{danijiar2018planet}, a purely model-based agent that learns the dynamics from interactions with the world. Their method predicts actions by fast online planning in latent space through images.
    \item \textit{Predictive Model}: We train a predictive model as proposed by Agrawal \etal~\cite{agrawal2016learning} that jointly learns a forward and inverse dynamics model for intuitive physics. The latent embedding is then used for MPC during planning.% We compare to their method for rope manipulation across a variety of tasks, including goal-directed planning and imitation from observations.
\end{itemize}

% Auto-encoder/PlaNet/Predictive Model/Ours(F)/Ours(FI)
%  \paragraph{CLOUD-RP (ours)} 
%  To understand the effect of our representation model, we build two modified models without representation part of the CLOUD, both state representation and action representation simultaneously and compare with the CLOUD qualitatively and quantitatively.

%  For the goal states of the manipulation, we designed various specific shape goals to test whether the agent could learn complex rope manipulation tasks such as moving it to the designed complex shapes in contrast of a straight state goal to test their performance on simple tasks.

% quantitative: consistent with other paper; ours best under noise, predictive model worse under noise than no noise; ours best under noise free; difference of ours under noise / w.o. noise smaller; FI > F

\begin{table}[t]
\centering
% \textbf{goal-directed planning}
\begin{tabular}{l cc cc}
\toprule
\multicolumn{5}{c}{Goal-directed planning (Success \%)} \\
\midrule
\multirow{2}{*}{Method} & \multicolumn{2}{c}{Deterministic environment} & %
    \multicolumn{2}{c}{Stochastic environment} \\
% \cline{2-5} 
\cmidrule(lr){2-3}\cmidrule(lr){4-5}
    & shaped & straight & shaped & straight \\
    \midrule
    Auto-encoder  & $32.8\%\pm{3.9\%} $ & $ 50.9\%\pm{2.9\%}$ & $16.5\%\pm{9.9\%}$ & $42.1\%\pm{7.5\%}$ \\
    PlaNet   & $29.8\%\pm{3.7\%}$ & $53.7\%\pm{3.2\%}$ & $19.9\%\pm{5.4\%}$ & $45.2\%\pm{6.1\%}$ \\ 
    Predictive Model & $34.8\%\pm{2.2\%}$ & $53.3\%\pm{0.7\%}$ & $18.5\%\pm{9.2\%}$ & $40.9\%\pm{7.3\%}$ \\ 
    \midrule
    CLOUD (F)    & $43.3\%\pm{1.5\%}$ & $53.4\%\pm{0.7\%}$ & $41.6\%\pm{2.3\%}$ & $\bm{54.6\%}\pm{\bm{1.2\%}}$ \\ 
    CLOUD (FI) & $\bm{49.9}\%\pm{\bm{1.4\%}}$ & $\bm{60.8\%}\pm{\bm{1.7\%}}$ & $\bm{43.2}\%\pm{\bm{1.5\%}}$ & $53.1\%\pm{6.7\%}$ \\ 
    \bottomrule
\end{tabular}
    \vspace{1em}
    \caption{\textbf{Success rate on the goal-directed planning task}. Our method outperforms all baseline methods. Prominently, it gets a negligible decline in performance in the stochastic environment, indicating that learning dynamics via contrastive estimation performs better and more robustly than deterministic models. }
    \label{tab:mpc}
\end{table}

Results show that our method consistently outperforms all baselines in both shaped goal state and straight goal state. As can be seen in Table~\ref{tab:mpc}, when the goal state is simpler, all approaches achieve better performance. We also show that training jointly with an inverse dynamics component instead of a single forward dynamics model performs better on rope manipulation. Such a jointly training strategy regularizes the state and action representation to extract useful information for planning. The poor performance of Auto-encoder also proves the importance of jointly optimizing forward and inverse dynamics models. 

Importantly, our method gets a relatively negligible decline in the success rate compared with all baselines under a stochastic environment. The most likely reason is that contrastive estimation assigns a low cost to all possible predictions instead of predicting the average future, given the fact that our method adopts the same architecture as the Predictive Model. Similarly, PlaNet uses a variational encoder, which leads to better performance under a stochastic environment. 

\begin{figure}[t]
\centering\includegraphics[scale=0.42]{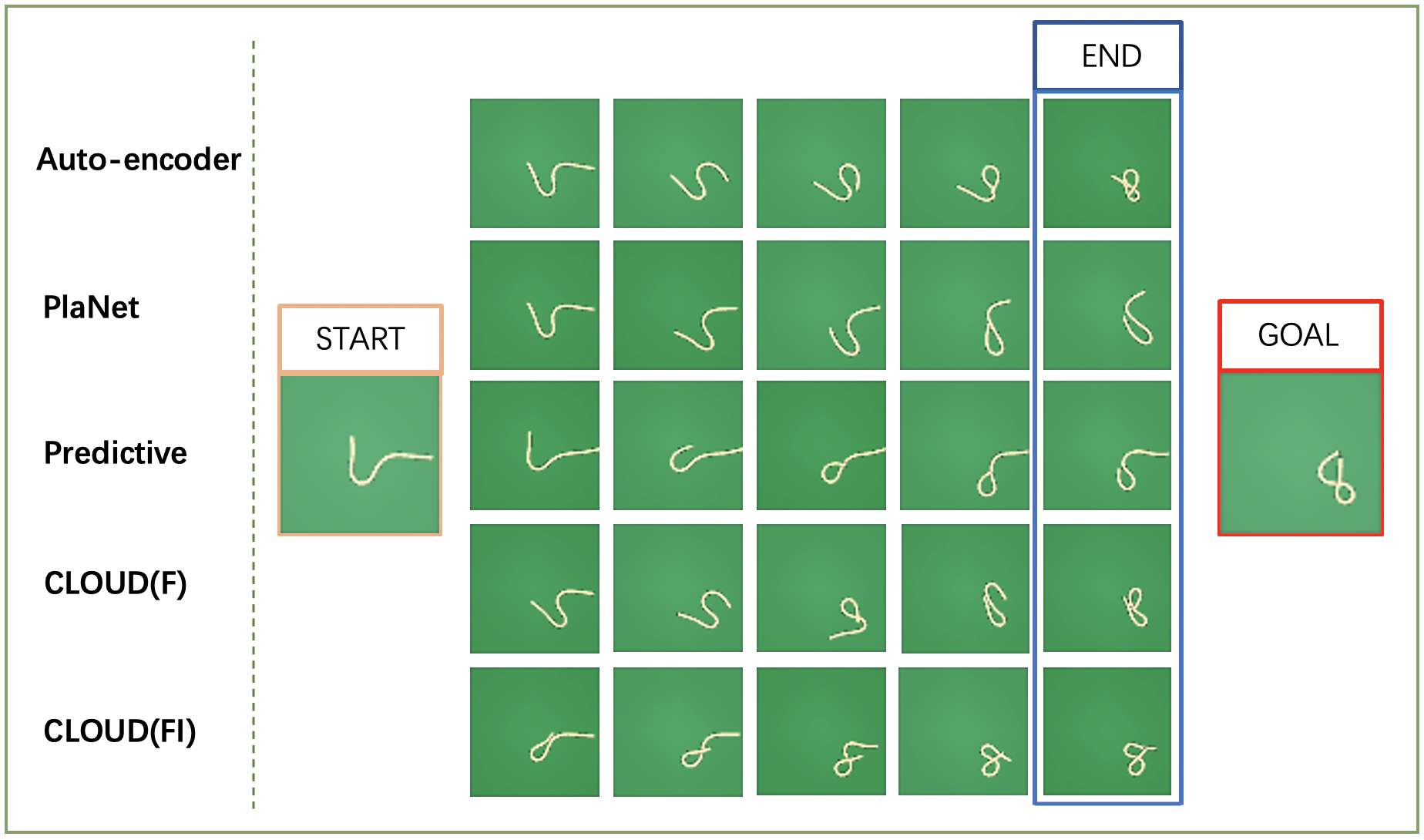}
  \caption{\textbf{Trajectories of different models on the goal-directed planning task.} All agents start from the same start state and are asked to reach the same goal state. Each trajectory is run for 20 actions. Note that the end states of our method achieves the highest similarity with goal state compare with all other methods.
%   \HZ{our result at the end state achieves the highest similarity, blabla...}
}
  \label{fig:mpc}
  \vspace{1em}
\end{figure}

We also show qualitative results of various approaches manipulating the rope under a stochastic environment in Figure~\ref{fig:mpc}. Each trajectory runs for 20 actions with the same start state and goal state. In the task of manipulating the rope into a "knot" shape, with our method, the agent successfully achieves the goal state. In comparison, all other methods fail to reach the goal state.

% quantitative: consistent with other paper; ours best under noise, predictive model worse under noise than no noise; ours best under noise free; difference of ours under noise / w.o. noise smaller; FI > F
% qualitative: reasonable

% \JW{Add Intermediate step, not only the final step}

% \subsubsection{Rope Manipulation}
%  We built our own custom environment for rope manipulation based on the DM Control API, which provides basic pipelines for simulation in reinforcement learning tasks. As the rope is deformable object, we utilize the composite objects from MuJoCo. Follow the experiment setup in [TODO: add Learning to Manipulate Deformable Objects without Demonstrations], we also train our models both on state and image observations. The state and observation part are the same as [] described, a fixed size of 3 color channel as RGB image and the locations of the composite objects. 
 
%  For the agent in the testing mode, the task here is to predict the action that could achieve the desired final state given the state and image-based representations collected from the simulation environment illustrated above.
 
% \subsubsection{Simulation}
% We can see that using our proposed method could achieve the desired goal successfully, and qualitatively speaking, much better performance than the baselines models. When the stochasticity is added to our simulation environment, from the table, we can see that there is a big drop down of success rate in the baseline models, while our model still remain the good performance. 

\subsection{Imitation from Observations}
% Dataset and Settings (Environment / Metric)
% Training Details (paragraph)
% Results: introduction of all baseline]

The goal of imitation from observation is to have the robot watch a sequence of images depicting each stage of the demonstration and then reproduce this demonstration on its own. We adopt the imitation method from Nair \etal~\cite{nair2017combining}. The robot receives a demonstration in the form of a sequence of images of the rope in intermediate states toward a final goal. We denote this sequence of demonstration as $\mathcal{D} = (d_0, d_1, ..., d_T)$, where $T$ is the length of demonstration. Let $s_1$ be the initial visual state of the robot and $d_i$ be the goal visual state. The robot first inputs the pair of states $(s_1, d_2)$ into the learned inverse dynamics model and executes the predicted action. Let $s_2$ be
the visual state of the world after the action is executed. The robot then inputs $(s_2, d_3)$ into the inverse model and executes the output action. This process is repeated iteratively for $T$ time steps.

\paragraph{Dataset and Settings.}
% \JW{explain here}
We use the same environment, dataset and metric as mentioned in Section~\ref{sec:data_goal} to quantitatively evaluate the performance of imitation from observations. In this task, the trajectory through which the agent achieves the final state is important. Therefore, we consider the average distance of the entire trajectory instead of only using final states when calculating the success rate (mean distance error threshold is set to 4 pixels of each geom).

\paragraph{Training Details.}
We use the same network architecture and training procedure as mentioned in Section~\ref{network}. We further use a 4-layer MLP to decode action representations back to actual actions during execution.

% During training of the inverse dynamics model, the states fed into the neural network are in format of neighbor images pair with corresponding actions. In detail, let $S = {D_t | t/in(1...T)}$ represent a demonstration sequence collected via random exploration, $A_1$ represent the initial image from agent's observation. At the beginning, the agent receives $(A_1, D_2)$ as the input pair of images into the inverse dynamics model and predicts the action. The agent apply the predicted action on the rope to transform from $A_1$ into $A_2$. Then the input pair becomes $(A_2, D_3)$, $(A_t-1, D_t)$ and so on. This process is repeated iteratively until t reaches T.

% We used approximately 150K pairs of neighbor images collected in a self-supervised manner via random exploration for training the inverse dynamics model. For fine-tuning period, another 10K neighbor images are used. As for the default setting, we use the learning rate of 1e-3 and weight decay of 1e-6 similar as in goal-directed planning task.

% Our proposed neural network architecture composed of four learnable components that are jointly trained via contrastive estimation, including a forward dynamics model, an inverse dynamics model, and two other representation model to encode high-dimensional images(or actions) into informative embeddings respectively. We ablate our forward dynamics and state representation component to quantitatively and qualitatively review their importance for our complete model in imitation from observations task.

\paragraph{Results.}

Table~\ref{tab:imi} shows the success rate of various approaches on imitation from observations. We evaluate two different versions of our method:

% Our goal-directed planning experiments aim to have the agent predict the next state given the current state and action. We compare our method with several classical baselines on the MPC task, including Auto-encoder, PlaNet, Predictive Model and a variant of our complete model. The details of all the models in comparison are described below. 

\begin{itemize}
    \item \textit{CLOUD (I)}: The method refers to a variant of CLOUD without using the forward dynamics model. The purpose of this variant is to particularly ablate the benefit of our forward dynamics model. 
    \item \textit{CLOUD (FI)}: In contrast to CLOUD (I), it refers to our complete model. During inference, we utilize its inverse dynamics component for imitation from observation.
\end{itemize}

We compare our results with the following baselines:

\begin{itemize}
   \item \textit{Nearest Neighbor baseline}: To evaluate whether the neural network simply memorizes the training data, we implement a nearest neighbor baseline. Given a pair of states $(s_t,d_{t+1})$, we find the state transition in the training set that is closest to $(s_t,d_{t+1})$. The action is then executed. We use Euclidean distance in the pixel space as the distance metric. 
    % In detail, given the current state $A_t$ from agent's observation and the next state $D_t+1$ from target demonstration, the agent first select the neighbor images $(M_k, M_k+1)$ which is the most similar pair as the pair $(A_t, D_t+1)$, then execute the corresponding action that transform the rope from $M_k$ into $M_k+1$ told by the memory bank.
    \item \textit{Self-supervised Imitation}: We also compare our method with self-supervised imitation proposed by Nair~\etal~\cite{nair2017combining}. We reimplement this baseline and train in our setup for a fair comparison.
\end{itemize}

\begin{table}[t]
\centering
\resizebox{\textwidth}{!}{ 
\begin{tabular}{l cc cc}
\toprule
\multicolumn{5}{c}{Imitation from observations (Success \%)} \\
\midrule
\multirow{2}{*}{Method} & \multicolumn{2}{c}{Deterministic environment} & %
    \multicolumn{2}{c}{Stochastic environment}\\
% \cline{2-5} 
\cmidrule(lr){2-3}\cmidrule(lr){4-5}
    & shaped & straight & shaped & straight \\
    \midrule
    Nearest Neighbor   & $17.5\%\pm{4.7\%}$ & $19.3\%\pm{3.8\%}$ & $10.7\%\pm{3.6\%}$ & $16.7\%\pm{5.1\%}$ \\ 
    Self-supervised Imitation & $20.6\%\pm{5.8\%}$ & $25.5\%\pm{2.4\%}$ & $12.3\%\pm{4.6\%}$ & $19.1\%\pm{2.9\%}$ \\ 
    \midrule
    CLOUD (I)    & $30.7\%\pm{4.5\%}$ & $\bm{56.8}\%\pm{\bm{3.9\%}}$ & $26.8\%\pm{3.7\%}$ & $48.1\%\pm{5.7\%}$ \\ 
    CLOUD (FI) & $\bm{36.7}\%\pm{\bm{2.8\%}}$ & $54.9\%\pm{2.3}$ & $\bm{32.4\%}\pm{\bm{3.1\%}}$ & $\bm{49.9\%}\pm{\bm{1.9\%}}$ \\ 
    \bottomrule
\end{tabular}
}
    \vspace{1em}
    \caption{\textbf{Success rate of imitation from observations.} The performance of our method is better than baseline methods both in straightening rope and manipulating rope in desired shapes. The gap between stochastic environment and deterministic environment shrinks via contrastive learning.
    % \HZ{this description is the same as Table 1.}
    }
    \label{tab:imi}
\end{table}

\begin{figure}[t]
% \textbf{Imitation Trajectory}\par\medskip
\centering\includegraphics[scale=0.38]{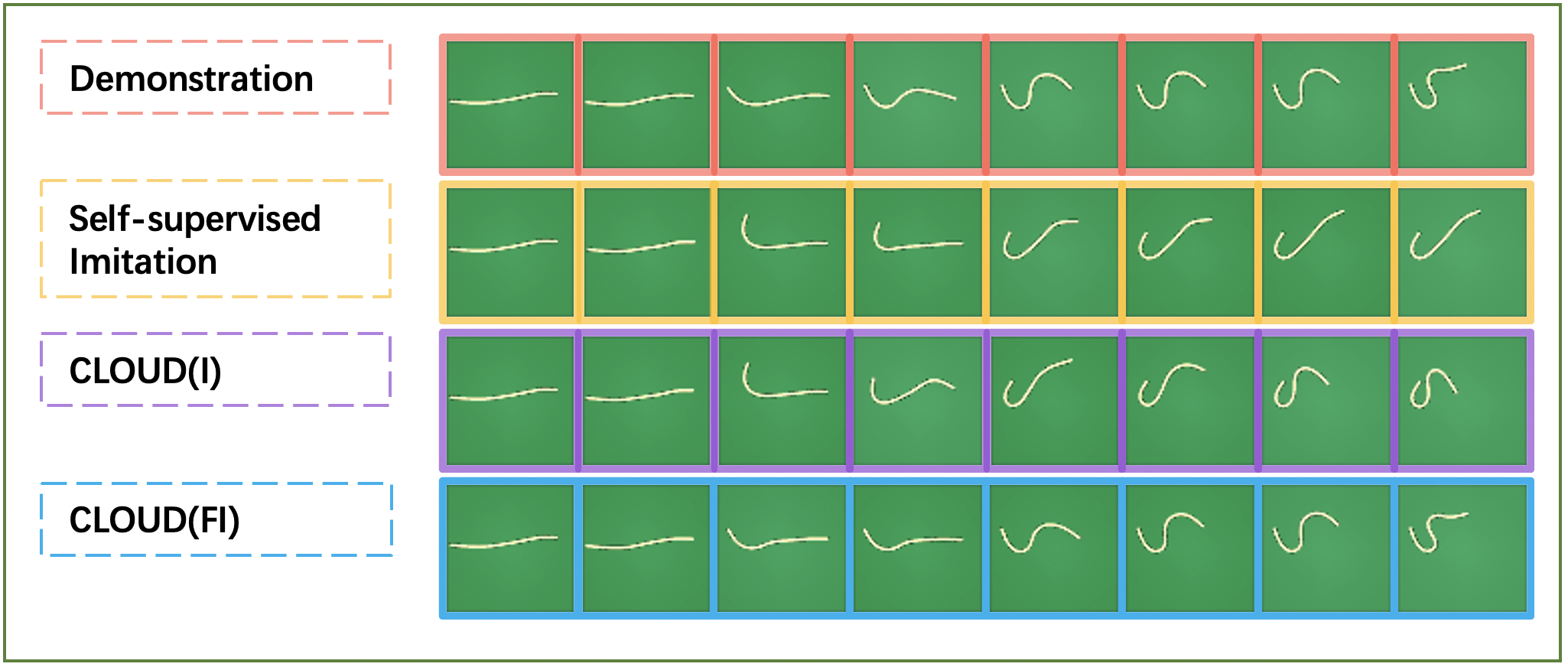}
  \caption{\textbf{Trajectories of imitation from observations.} The upper row represents an example expert demonstration. The following rows show the states achieve by the agent during imitation. The trajectory generated by our method achieves the highest similarity with the demonstration.
%   \HZ{incomplete?}
  }
  \label{fig:imitraj}
\end{figure}
% Our imitation from observations experiment aim to have the agent produce action sequences through trajectory provided by an expert generated demonstration both in deterministic and stochastic environment, without any access to expert actions during training or inference. The following baselines and a variant of our complete model will be evaluated and compared to in this paragraph:

% \begin{itemize}
    
%     \item \textit{CLOUD(I)}: The purpose of this variant is to particularly ablate the benefit of our forward dynamics model with respect to just having inverse dynamics model for imitation from observations.
%     \item \textit{CLOUD(FI)}:  In contrast to CLOUD(I), it refers to our complete method without any modification as a baseline model. During inference, we utilize its inverse dynamics component for imitation from observations.
% \end{itemize}

Results show that our method outperforms all baselines methods for imitation from observations under both shaped and straight goal states. It worth noticing that Self-supervised Imitation performs similarly with Nearest Neighbor baseline, especially under a stochastic environment, which suggests that deterministic prediction somehow memorize the training data and fail to generalize to the stochastic environments. On the contrary, our methods perform much better than the baseline methods, which further proves the importance of contrastive estimation.

Comparing CLOUD (I) with CLOUD (FI), we show that the state and action representation model can benefit from jointly optimizing forward and inverse dynamics model. The reason is the joint dynamics model regularizes state and action representations to only the information relevant to dynamics. By limiting the dimension of these representations, irrelevant information (e.g. background color, lighting) can be filtered automatically, which leads to better generalization.

Figure~\ref{fig:imitraj} qualitatively shows that our method is capable of imitating given demonstrations. We see that our method more accurately re-configure the rope to imitate the demonstration comparing with baseline methods.

% \HZ{some more discussions here.}

% \subsection{Model Comparisons}
% The left part in Figure~\ref{fig:modcom} shows the average distance error of the models in comparison, which indicates the best performance of our model from the deviation metric perspective. The training process of CLOUD and the other two baseline models Visual Control and Forward-Inverse is shown in the right part of Figure~\ref{fig:modcom}, which indicates that our proposed CLOUD model could be trained effectively compared to the baselines. 

% \begin{figure}[!htb]
% \centering
% \minipage{0.5\textwidth}
%   \includegraphics[width=\linewidth]{figures/plot/mpcDistancex.png}
% \endminipage\hfill
% \minipage{0.5\textwidth}
%   \includegraphics[width=\linewidth]{figures/plot/Training-Process.png}
% \endminipage\hfill
% \caption{Model Comparison}
% \label{fig:modcom}
% \end{figure}

% MPC
% Auto-encoder/PlaNet/Predictive Model/Ours(F)/Ours(FI)
% Qualitative / Quantitative

% quantitative: consistent with other paper; ours best under noise, predictive model worse under noise than no noise; ours best under noise free; difference of ours under noise / w.o. noise smaller; FI > F
% qualitative: reasonable

\section{Conclusion}

In conclusion, we proposed CLOUD, a contrastive estimation framework for unsupervised dynamics learning. We show that our learned dynamics are effective for both goal conditioned planning and imitation from observations. Although we only evaluate our method on rope manipulation, there are no task specific assumptions. In future work, we plan to extend CLOUD on more robotic tasks under more complex environments.
We hope our work points a new way to learn plannable representations, dynamics models that can handle stochasticity of the environment. 
% \HZ{In future work, we plan to extend CLOUD on more robotic tasks.}

% \input{supplementary.tex}

% \clearpage

% The acknowledgments are automatically included only in the final version of the paper.
\acknowledgments{If a paper is accepted, the final camera-ready version will (and probably should) include acknowledgments. All acknowledgments go at the end of the paper, including thanks to reviewers who gave useful comments, to colleagues who contributed to the ideas, and to funding agencies and corporate sponsors that provided financial support.}

%===============================================================================

{\small \bibliography{reference}}  % .bib

\begin{thebibliography}{48}
\providecommand{\natexlab}[1]{#1}
\providecommand{\url}[1]{\texttt{#1}}
\expandafter\ifx\csname urlstyle\endcsname\relax
  \providecommand{\doi}[1]{doi: #1}\else
  \providecommand{\doi}{doi: \begingroup \urlstyle{rm}\Url}\fi

\bibitem[De~Schutter and Van~Brussel(1988)]{de1988compliant}
J.~De~Schutter and H.~Van~Brussel.
\newblock Compliant robot motion i. a formalism for specifying compliant motion
  tasks.
\newblock \emph{IJRS}, 7\penalty0 (4):\penalty0 3--17, 1988.

\bibitem[Karayiannidis et~al.(2016)Karayiannidis, Smith, Barrientos, {\"O}gren,
  and Kragic]{karayiannidis2016adaptive}
Y.~Karayiannidis, C.~Smith, F.~E.~V. Barrientos, P.~{\"O}gren, and D.~Kragic.
\newblock An adaptive control approach for opening doors and drawers under
  uncertainties.
\newblock \emph{IEEE Transactions on Robotics}, 32\penalty0 (1):\penalty0
  161--175, 2016.

\bibitem[Kurutach et~al.(2018)Kurutach, Tamar, Yang, Russell, and
  Abbeel]{kurutach2018learning}
T.~Kurutach, A.~Tamar, G.~Yang, S.~J. Russell, and P.~Abbeel.
\newblock Learning plannable representations with causal infogan.
\newblock In \emph{NeurIPS}, pages 8733--8744, 2018.

\bibitem[Hafner et~al.(2019)Hafner, Lillicrap, Fischer, Villegas, Ha, Lee, and
  Davidson]{hafner2019learning}
D.~Hafner, T.~Lillicrap, I.~Fischer, R.~Villegas, D.~Ha, H.~Lee, and
  J.~Davidson.
\newblock Learning latent dynamics for planning from pixels.
\newblock In \emph{ICML}, pages 2555--2565, 2019.

\bibitem[Pathak et~al.(2018)Pathak, Mahmoudieh, Luo, Agrawal, Chen, Shentu,
  Shelhamer, Malik, Efros, and Darrell]{pathakICLR18zeroshot}
D.~Pathak, P.~Mahmoudieh, G.~Luo, P.~Agrawal, D.~Chen, Y.~Shentu, E.~Shelhamer,
  J.~Malik, A.~A. Efros, and T.~Darrell.
\newblock Zero-shot visual imitation.
\newblock In \emph{ICLR}, 2018.

\bibitem[{Nagabandi} et~al.(2018){Nagabandi}, {Kahn}, {Fearing}, and
  {Levine}]{8463189}
A.~{Nagabandi}, G.~{Kahn}, R.~S. {Fearing}, and S.~{Levine}.
\newblock Neural network dynamics for model-based deep reinforcement learning
  with model-free fine-tuning.
\newblock In \emph{ICRA}, pages 7559--7566, 2018.

\bibitem[Kaiser et~al.(2019)Kaiser, Babaeizadeh, Mi{\l}os, Osi{\'n}ski,
  Campbell, Czechowski, Erhan, Finn, Kozakowski, Levine,
  et~al.]{kaiser2019model}
{\L}.~Kaiser, M.~Babaeizadeh, P.~Mi{\l}os, B.~Osi{\'n}ski, R.~H. Campbell,
  K.~Czechowski, D.~Erhan, C.~Finn, P.~Kozakowski, S.~Levine, et~al.
\newblock Model based reinforcement learning for atari.
\newblock In \emph{ICLR}, 2019.

\bibitem[Nair et~al.(2017)Nair, Chen, Agrawal, Isola, Abbeel, Malik, and
  Levine]{nair2017combining}
A.~Nair, D.~Chen, P.~Agrawal, P.~Isola, P.~Abbeel, J.~Malik, and S.~Levine.
\newblock Combining self-supervised learning and imitation for vision-based
  rope manipulation.
\newblock In \emph{ICRA}, pages 2146--2153. IEEE, 2017.

\bibitem[Agrawal et~al.(2016)Agrawal, Nair, Abbeel, Malik, and
  Levine]{agrawal2016learning}
P.~Agrawal, A.~V. Nair, P.~Abbeel, J.~Malik, and S.~Levine.
\newblock Learning to poke by poking: Experiential learning of intuitive
  physics.
\newblock In \emph{NeurIPS}, pages 5074--5082, 2016.

\bibitem[Kumar et~al.(2016)Kumar, Todorov, and Levine]{kumar2016optimal}
V.~Kumar, E.~Todorov, and S.~Levine.
\newblock Optimal control with learned local models: Application to dexterous
  manipulation.
\newblock In \emph{ICRA}, pages 378--383. IEEE, 2016.

\bibitem[Lake et~al.(2017)Lake, Ullman, Tenenbaum, and
  Gershman]{lake2017building}
B.~M. Lake, T.~D. Ullman, J.~B. Tenenbaum, and S.~J. Gershman.
\newblock Building machines that learn and think like people.
\newblock \emph{Behavioral and brain sciences}, 40, 2017.

\bibitem[Laskin et~al.(2020)Laskin, Srinivas, and
  Abbeel]{laskin_srinivas2020curl}
M.~Laskin, A.~Srinivas, and P.~Abbeel.
\newblock Curl: Contrastive unsupervised representations for reinforcement
  learning.
\newblock \emph{ICML}, 2020.

\bibitem[Chen et~al.(2016)Chen, Duan, Houthooft, Schulman, Sutskever, and
  Abbeel]{chen2016infogan}
X.~Chen, Y.~Duan, R.~Houthooft, J.~Schulman, I.~Sutskever, and P.~Abbeel.
\newblock Infogan: Interpretable representation learning by information
  maximizing generative adversarial nets.
\newblock In \emph{NeurIPS}, pages 2172--2180, 2016.

\bibitem[Ebert et~al.(2018)Ebert, Finn, Dasari, Xie, Lee, and
  Levine]{ebert2018visual}
F.~Ebert, C.~Finn, S.~Dasari, A.~Xie, A.~Lee, and S.~Levine.
\newblock Visual foresight: Model-based deep reinforcement learning for
  vision-based robotic control.
\newblock \emph{arXiv preprint arXiv:1812.00568}, 2018.

\bibitem[Pathak et~al.(2017)Pathak, Agrawal, Efros, and
  Darrell]{pathak2017curiosity}
D.~Pathak, P.~Agrawal, A.~A. Efros, and T.~Darrell.
\newblock Curiosity-driven exploration by self-supervised prediction.
\newblock In \emph{ICML}, pages 2778--2787, 2017.

\bibitem[Chua et~al.(2018)Chua, Calandra, McAllister, and Levine]{chua2018deep}
K.~Chua, R.~Calandra, R.~McAllister, and S.~Levine.
\newblock Deep reinforcement learning in a handful of trials using
  probabilistic dynamics models.
\newblock In \emph{NeurIPS}, pages 4754--4765, 2018.

\bibitem[Houthooft et~al.(2016)Houthooft, Chen, Duan, Schulman, De~Turck, and
  Abbeel]{houthooft2016vime}
R.~Houthooft, X.~Chen, Y.~Duan, J.~Schulman, F.~De~Turck, and P.~Abbeel.
\newblock Vime: Variational information maximizing exploration.
\newblock In \emph{NeurIPS}, pages 1109--1117, 2016.

\bibitem[Chen et~al.(2020)Chen, Kornblith, Norouzi, and Hinton]{chen2020simple}
T.~Chen, S.~Kornblith, M.~Norouzi, and G.~Hinton.
\newblock A simple framework for contrastive learning of visual
  representations.
\newblock \emph{arXiv preprint arXiv:2002.05709}, 2020.

\bibitem[LeCun et~al.(2006)LeCun, Chopra, Hadsell, Ranzato, and
  Huang]{lecun2006tutorial}
Y.~LeCun, S.~Chopra, R.~Hadsell, M.~Ranzato, and F.~Huang.
\newblock A tutorial on energy-based learning.
\newblock \emph{Predicting structured data}, 1\penalty0 (0), 2006.

\bibitem[Chandak et~al.(2019)Chandak, Theocharous, Kostas, Jordan, and
  Thomas]{chandak2019learning}
Y.~Chandak, G.~Theocharous, J.~Kostas, S.~Jordan, and P.~Thomas.
\newblock Learning action representations for reinforcement learning.
\newblock In \emph{ICML}, pages 941--950, 2019.

\bibitem[Hadsell et~al.(2006)Hadsell, Chopra, and
  LeCun]{hadsell2006dimensionality}
R.~Hadsell, S.~Chopra, and Y.~LeCun.
\newblock Dimensionality reduction by learning an invariant mapping.
\newblock In \emph{CVPR}, volume~2, pages 1735--1742. IEEE, 2006.

\bibitem[Wu et~al.(2018)Wu, Xiong, Yu, and Lin]{wu2018unsupervised}
Z.~Wu, Y.~Xiong, S.~X. Yu, and D.~Lin.
\newblock Unsupervised feature learning via non-parametric instance
  discrimination.
\newblock In \emph{CVPR}, pages 3733--3742, 2018.

\bibitem[Ye et~al.(2019)Ye, Zhang, Yuen, and Chang]{ye2019unsupervised}
M.~Ye, X.~Zhang, P.~C. Yuen, and S.-F. Chang.
\newblock Unsupervised embedding learning via invariant and spreading instance
  feature.
\newblock In \emph{CVPR}, pages 6210--6219, 2019.

\bibitem[Tian et~al.(2019)Tian, Krishnan, and Isola]{tian2019contrastive}
Y.~Tian, D.~Krishnan, and P.~Isola.
\newblock Contrastive multiview coding.
\newblock \emph{arXiv preprint arXiv:1906.05849}, 2019.

\bibitem[Doersch and Zisserman(2017)]{doersch2017multi}
C.~Doersch and A.~Zisserman.
\newblock Multi-task self-supervised visual learning.
\newblock In \emph{ICCV}, 2017.

\bibitem[Ji et~al.(2019)Ji, Henriques, and Vedaldi]{ji2019invariant}
X.~Ji, J.~F. Henriques, and A.~Vedaldi.
\newblock Invariant information clustering for unsupervised image
  classification and segmentation.
\newblock In \emph{ICCV}, pages 9865--9874, 2019.

\bibitem[He et~al.(2020)He, Fan, Wu, Xie, and Girshick]{he2020momentum}
K.~He, H.~Fan, Y.~Wu, S.~Xie, and R.~Girshick.
\newblock Momentum contrast for unsupervised visual representation learning.
\newblock In \emph{CVPR}, pages 9729--9738, 2020.

\bibitem[Chen et~al.(2020)Chen, Fan, Girshick, and He]{chen2020mocov2}
X.~Chen, H.~Fan, R.~Girshick, and K.~He.
\newblock Improved baselines with momentum contrastive learning.
\newblock \emph{arXiv preprint arXiv:2003.04297}, 2020.

\bibitem[Wu et~al.(2015)Wu, Yildirim, Lim, Freeman, and
  Tenenbaum]{wu2015galileo}
J.~Wu, I.~Yildirim, J.~J. Lim, B.~Freeman, and J.~Tenenbaum.
\newblock Galileo: Perceiving physical object properties by integrating a
  physics engine with deep learning.
\newblock In \emph{NeurIPS}, pages 127--135, 2015.

\bibitem[Wu et~al.(2017)Wu, Lu, Kohli, Freeman, and Tenenbaum]{wu2017learning}
J.~Wu, E.~Lu, P.~Kohli, B.~Freeman, and J.~Tenenbaum.
\newblock Learning to see physics via visual de-animation.
\newblock In \emph{NeurIPS}, pages 153--164, 2017.

\bibitem[Watters et~al.(2017)Watters, Zoran, Weber, Battaglia, Pascanu, and
  Tacchetti]{watters2017visual}
N.~Watters, D.~Zoran, T.~Weber, P.~Battaglia, R.~Pascanu, and A.~Tacchetti.
\newblock Visual interaction networks: Learning a physics simulator from video.
\newblock In \emph{NeurIPS}, pages 4539--4547, 2017.

\bibitem[Ye et~al.(2018)Ye, Wang, Davidson, and Gupta]{ye2018interpretable}
T.~Ye, X.~Wang, J.~Davidson, and A.~Gupta.
\newblock Interpretable intuitive physics model.
\newblock In \emph{ECCV}, pages 87--102, 2018.

\bibitem[Ehrhardt et~al.(2019)Ehrhardt, Monszpart, Mitra, and
  Vedaldi]{ehrhardt2019taking}
S.~Ehrhardt, A.~Monszpart, N.~J. Mitra, and A.~Vedaldi.
\newblock Taking visual motion prediction to new heightfields.
\newblock \emph{CVIU}, 181:\penalty0 14--25, 2019.

\bibitem[Fragkiadaki et~al.(2015)Fragkiadaki, Agrawal, Levine, and
  Malik]{fragkiadaki2015learning}
K.~Fragkiadaki, P.~Agrawal, S.~Levine, and J.~Malik.
\newblock Learning visual predictive models of physics for playing billiards.
\newblock \emph{arXiv preprint arXiv:1511.07404}, 2015.

\bibitem[Byravan and Fox(2017)]{byravan2017se3}
A.~Byravan and D.~Fox.
\newblock Se3-nets: Learning rigid body motion using deep neural networks.
\newblock In \emph{ICRA}, pages 173--180. IEEE, 2017.

\bibitem[{Pinto} and {Gupta}(2017)]{7989249}
L.~{Pinto} and A.~{Gupta}.
\newblock Learning to push by grasping: Using multiple tasks for effective
  learning.
\newblock In \emph{ICRA}, pages 2161--2168, 2017.

\bibitem[Yan et~al.(2020)Yan, Vangipuram, Abbeel, and Pinto]{yan2020learning}
W.~Yan, A.~Vangipuram, P.~Abbeel, and L.~Pinto.
\newblock Learning predictive representations for deformable objects using
  contrastive estimation.
\newblock \emph{Conference on Robot Learning}, 2020.

\bibitem[Edwards et~al.(2019)Edwards, Sahni, Schroecker, and
  Isbell]{edwards2019imitating}
A.~Edwards, H.~Sahni, Y.~Schroecker, and C.~Isbell.
\newblock Imitating latent policies from observation.
\newblock In \emph{ICML}, pages 1755--1763, 2019.

\bibitem[Liu et~al.(2018)Liu, Gupta, Abbeel, and Levine]{liu2018imitation}
Y.~Liu, A.~Gupta, P.~Abbeel, and S.~Levine.
\newblock Imitation from observation: Learning to imitate behaviors from raw
  video via context translation.
\newblock In \emph{ICRA}, pages 1118--1125. IEEE, 2018.

\bibitem[Ho and Ermon(2016)]{ho2016generative}
J.~Ho and S.~Ermon.
\newblock Generative adversarial imitation learning.
\newblock In \emph{NeurIPS}, pages 4565--4573, 2016.

\bibitem[Torabi et~al.(2018)Torabi, Warnell, and Stone]{torabi2018behavioral}
F.~Torabi, G.~Warnell, and P.~Stone.
\newblock Behavioral cloning from observation.
\newblock In \emph{IJCAI}, pages 4950--4957, 2018.

\bibitem[Maaten and Hinton(2008)]{maaten2008visualizing}
L.~v.~d. Maaten and G.~Hinton.
\newblock Visualizing data using t-sne.
\newblock \emph{Journal of machine learning research}, 9\penalty0
  (Nov):\penalty0 2579--2605, 2008.

\bibitem[Klingemann(2015)]{mario2015RasterFairy}
M.~Klingemann.
\newblock Rasterfairy, 2015.
\newblock \url{https://github.com/Quasimondo/RasterFairy}.

\bibitem[Tassa et~al.(2018)Tassa, Doron, Muldal, Erez, Li, Casas, Budden,
  Abdolmaleki, Merel, Lefrancq, et~al.]{tassa2018deepmind}
Y.~Tassa, Y.~Doron, A.~Muldal, T.~Erez, Y.~Li, D.~d.~L. Casas, D.~Budden,
  A.~Abdolmaleki, J.~Merel, A.~Lefrancq, et~al.
\newblock Deepmind control suite.
\newblock \emph{arXiv preprint arXiv:1801.00690}, 2018.

\bibitem[Todorov et~al.(2012)Todorov, Erez, and Tassa]{todorov2012mujoco}
E.~Todorov, T.~Erez, and Y.~Tassa.
\newblock Mujoco: A physics engine for model-based control.
\newblock In \emph{2012 IEEE/RSJ International Conference on Intelligent Robots
  and Systems}, pages 5026--5033. IEEE, 2012.

\bibitem[Kingma and Ba(2014)]{kingma2014adam}
D.~P. Kingma and J.~Ba.
\newblock Adam: A method for stochastic optimization.
\newblock \emph{arXiv preprint arXiv:1412.6980}, 2014.

\bibitem[{Lange} and {Riedmiller}(2010)]{lange2010autoencoder}
S.~{Lange} and M.~{Riedmiller}.
\newblock Deep auto-encoder neural networks in reinforcement learning.
\newblock In \emph{IJCNN}, pages 1--8, 2010.

\bibitem[Hafner et~al.(2018)Hafner, Lillicrap, Fischer, Villegas, Ha, Lee, and
  Davidson]{danijiar2018planet}
D.~Hafner, T.~P. Lillicrap, I.~Fischer, R.~Villegas, D.~Ha, H.~Lee, and
  J.~Davidson.
\newblock Learning latent dynamics for planning from pixels.
\newblock \emph{CoRR}, abs/1811.04551, 2018.
\newblock URL \url{http://arxiv.org/abs/1811.04551}.

\end{thebibliography}

\end{document}